\documentclass{article}

\PassOptionsToPackage{numbers, compress}{natbib}

\usepackage[preprint]{neurips_2023}

\usepackage[utf8]{inputenc} 
\usepackage[T1]{fontenc}    
\usepackage{hyperref}       
\usepackage{url}            
\usepackage{booktabs}       
\usepackage{amsfonts}       
\usepackage{nicefrac}       
\usepackage{microtype}      
\usepackage{xcolor}         
\usepackage{graphicx}
\usepackage{titlesec}
\titleformat{\subsubsection}[runin]{\bfseries}{}{}{}[]

\title{Evaluating Contextual Intelligence in Recyclability: \\ A Comprehensive Study of Image-Based Reasoning Systems}
\author{
  Eliot Park \\
  Harvard College\\
  Cambridge, MA 02138 \\
  \texttt{eliot\_park@college.harvard.edu} \\
  \And
  Abhi Kumar \\
  Stanford University \\
  Stanford, CA 94305 \\
  \texttt{abhi1@stanford.edu} \\
  \And
  Pranav Rajpurkar \\
  Department of Biomedical Informatics\\
  Harvard Medical School\\
  Boston, MA 02115 \\
  \texttt{pranav\_rajpurkar@hms.harvard.edu} \\
}

\begin{document}
\maketitle
\begin{abstract}
While the importance of efficient recycling is widely acknowledged, accurately determining the recyclability of items and their proper disposal remains a complex task for the general public. In this study, we explore the application of cutting-edge vision-language models (GPT-4o, GPT-4o-mini, and Claude 3.5) for predicting the recyclability of commonly disposed items. Utilizing a curated dataset of images, we evaluated the models’ ability to match objects to appropriate recycling bins, including assessing whether the items could physically fit into the available bins. Additionally, we investigated the models’ performance across several challenging scenarios: (i) adjusting predictions based on location-specific recycling guidelines; (ii) accounting for contamination or structural damage; and (iii) handling objects composed of multiple materials. Our findings highlight the significant advancements in contextual understanding offered by these models compared to previous iterations, while also identifying areas where they still fall short. The continued refinement of context-aware models is crucial for enhancing public recycling practices and advancing environmental sustainability.
\end{abstract}

\section{Introduction}

Effective waste management, particularly through recycling, is essential in promoting environmental sustainability. In 2018, the United States generated approximately 292.4 million tons of municipal solid waste, equating to 4.9 pounds per person per day \citep{epa-website}. Of this, 32.1 percent was either recycled or composted, a notable achievement but one that highlights the significant proportion of waste still destined for landfills. Within this waste stream, certain materials, such as paper and paperboard, achieved recycling rates as high as 68.2 percent, while others, such as plastics, lagged far behind at just 8.7 percent \citep{epa-website}. These disparities underscore the need for innovative approaches to improve recycling rates across all categories, including a better understanding by the general public in distinguishing which items should be recycled.

\section{Related Works}

In a recent work [2], the potential of general vision-language models, specifically Contrastive Language-Image Pretraining (CLIP) \cite{CLIP}, for automating the classification of waste materials for recycling was explored. The results were substantially better compared to previous approaches using simple convolution neural networks \cite{YangThung,LiuLiu,MaoLin} with the model achieving an accuracy of 89\% in zero-shot classification into a dozen different disposal methods. However, the approach had notable limitations. CLIP’s reliance on a predefined list of potential items meant it struggled with items outside of this list, reducing its effectiveness in real-world applications where waste items are highly varied. In particular, the model was not designed to handle common but challenging cases such as greasy, dirty, or broken items, which often complicate the recycling process.
The current study seeks to address these shortcomings by utilizing state-of-the-art models GPT-4o \cite{gpt4}, GPT-4o-mini, and Claude 3.5 \cite{claude}, which are all equipped with advanced vision capabilities. These models enhance the system’s ability to classify a broader range of waste items, including those with unique characteristics that significantly impact their recyclability, such as contamination and structural damage. By integrating more diverse and representative data, and leveraging these cutting-edge models, we aim to develop a more accurate and robust tool for waste classification, ultimately contributing to more efficient recycling practices.

\begin{figure}
  \centering
  \includegraphics[width=5.5in]{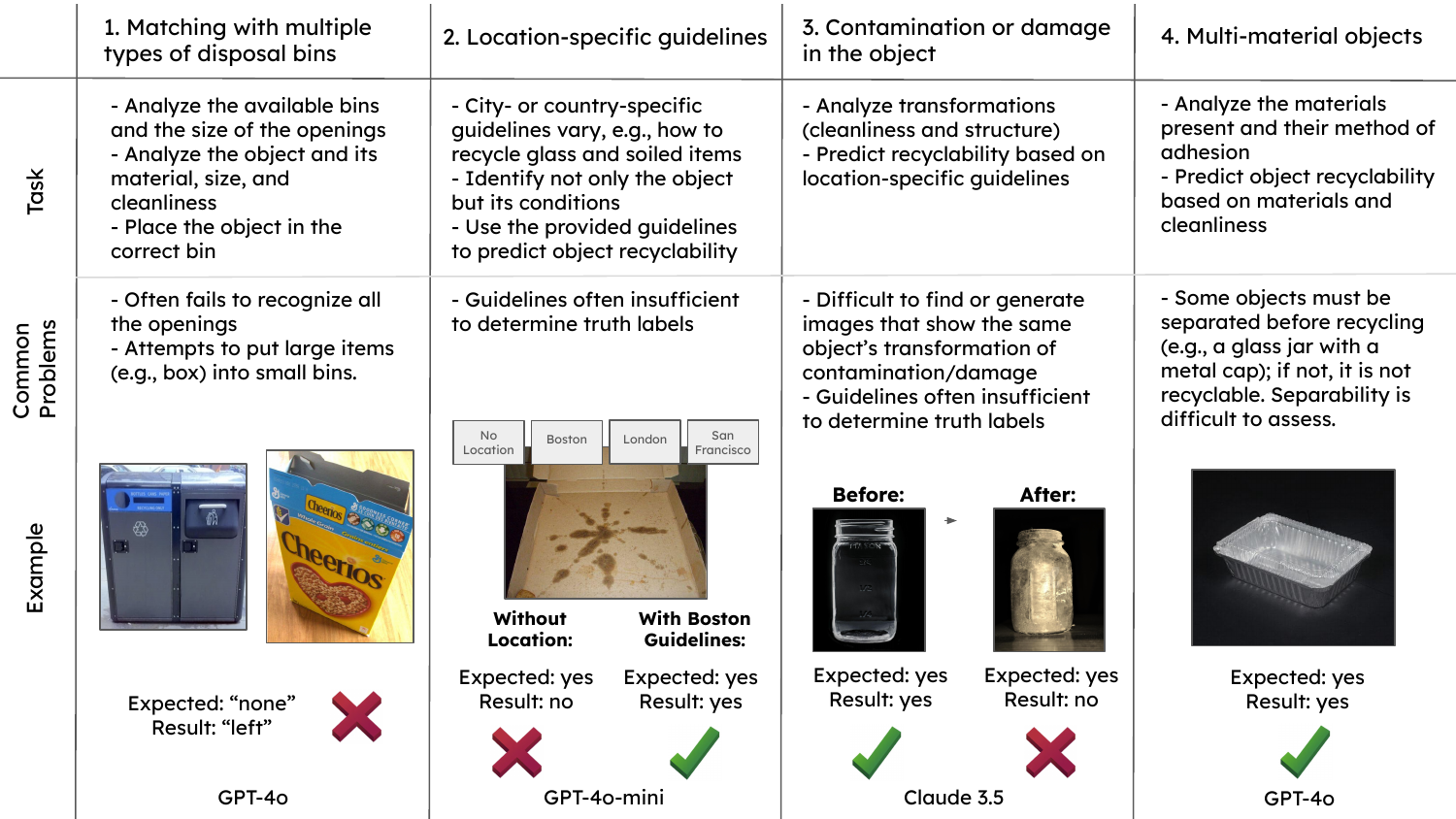}
  \caption{Overview of our study. Four contextual predictions are tested for three models. }
  \label{bin}
\end{figure}

\section{Methodology}

The overview of the experiments are shown in Figure 1. We collected images through a combination of three methods: Google Images, DALL·E image generation, and personal photography. The goal was to compile a diverse and representative database of commonly disposed items, ensuring that the images were clear, realistic, and contextually appropriate. We generated an initial list of 100 commonly disposed items using ChatGPT, which provided a broad array of materials across various categories. After refining this list, we manually gathered images for each item, focusing on capturing realistic depictions rather than stock photos with blank backgrounds.

The materials included in the database covered a wide spectrum, with a focus on frequently encountered waste types: Cardboard, Electronics, Glass, Plastic, Metal, Organics, Paper, Styrofoam, Textiles, and Wood. Each category was represented by 10 images, yielding a total of 100 images.

Three models were used: \textit{(i)} GPT-4o (released May 2024): a multimodal (text, images, audio) generative pre-trained transformer from OpenAI \cite{gpt4}. \textit{(ii)} GPT-4o-mini (July 2024): a small and less expensive version of GPT-4o. \textit{(iii)} Claude 3.5 Sonnet (Jun 2024): a generative pre-trained transformer from Anthropic.

\subsection{Experiment 1: Predicting recyclability based on recognizing different types of waste bin}
One challenge in a real-life situation is the different types of recycling options that are available at the time of disposal. In this experiment, we aimed to evaluate the model’s ability to classify waste items based on their compatibility with different types of disposal bins. Each API call involved two images: one of a bin and one of an item from the previously curated database. The models were prompted to consider both the size of the bin opening and the physical characteristics of the item, such as shape, size, and material, to determine the correct disposal method.

The experiment utilized three distinct bin images to represent common disposal scenarios: \textit{(i) BigBelly Bins:} These are urban bins found in many cities, featuring two categories (trash and recycling) with small openings designed to limit contamination. A major challenge for the models was to infer the size of the opening compared to the size of the item. \textit{(ii) Residential Bins:} Standard 96-gallon residential bins typically used for household waste collection. \textit{(iii) Three-Category Urban Bins:} These include categories for trash, recycling, and compost, commonly found in urban settings.  Each of the 100 images was tested against all three bin images, resulting in a comprehensive evaluation of the model’s ability to accurately classify items across varying disposal contexts. 



\subsection{Experiment 2: Predicting recyclability using location-specific guidelines}
Another complication is the differing recyclability guidelines depending on the city and country. This is a complex issue driven by the specific machines used at each recycling collection center and which distributor purchases the recycled items for a city. Thus, this experiment was designed to assess the model’s ability to adapt its waste classification based on location-specific recycling guidelines. We used the same database of 100 items across four tests, each incorporating varying levels of guideline specificity to evaluate how well the model could adjust to different local recycling practices.

The tests included: No Guidelines (165 word prompt) as the baseline and \textit{(i)} Boston Guidelines (377 word prompt); \textit{(ii)} London Guidelines (499 word prompt); \textit{(iii)} San Francisco Guidelines (557 word prompt). Across these tests, five items displayed a change in recyclability status when transitioning from Boston’s guidelines to those of London and San Francisco. This variation underscores the importance of local context in determining the correct disposal method, as items deemed recyclable in one city may not be accepted in another. 

\subsection{Experiment 3: Predicting Recyclability Based on Phase Changes}
Here, we aimed to evaluate the model’s ability to classify waste items based on transformations that may affect their recyclability. The experiment utilized a set of 40 images, each labeled as either “before” or “after” to depict the same object in two different states: pre-transformation (clean) and post-transformation. The transformations were divided into two categories: \textit{(i) Contamination}, with 20 images displayed objects before and after contamination, such as food residue or liquid spills; \textit{(ii) Structural change}, with 20 images captured objects before and after structural changes, including damage like tearing, breaking, or crumpling.


For each pair, the model was presented with both images simultaneously and tasked with assessing the transformation that occurred to determine the correct recyclability status. This experiment’s design provided a comprehensive evaluation of the model’s ability to consider physical changes that can affect whether an item should be recycled or discarded. The focus on 40 image pairs allowed for a detailed analysis of the model’s performance across both contamination and structural changes, providing insights into its ability to adapt to real-world scenarios where the condition of an item plays a crucial role in determining its recyclability.

\subsection{Experiment 4: Predicting Recyclability of Multi-Material Objects}

This experiment assessed the model’s ability to classify items made of multiple materials. A set of 50 images was used, each representing an object composed of more than one material, such as an aluminum takeout tray with a detachable plastic top or a glass jar with a metal lid. These images were passed through the model individually to evaluate its effectiveness in determining the recyclability of complex, multi-material objects.

\section{Results and Analysis}

\begin{figure}
  \centering
  \includegraphics[width=5.5in]{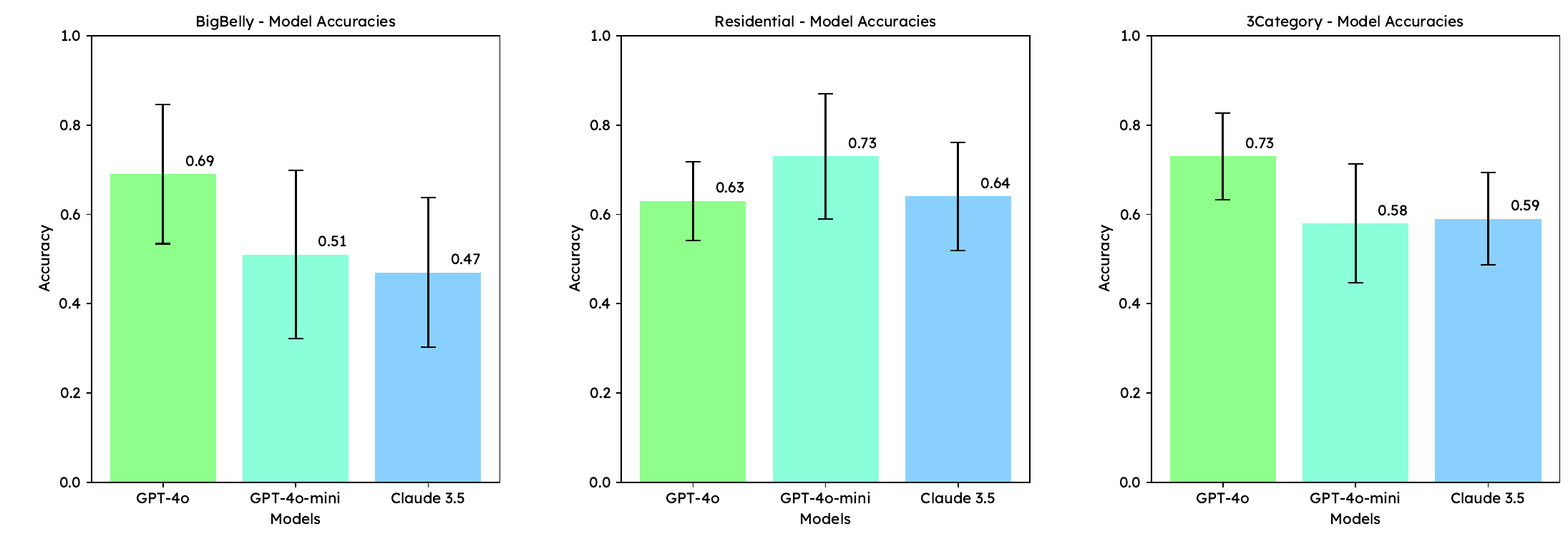}
  \caption{Performance on Bin Testing (Experiment 1). Recyclability prediction was made in the context of the available bin type, requiring accurate assessment of size of the item and the opening.}
  \label{bin}
\end{figure}

\subsection{Experiment 1: Bin Testing (Figure 2)}

\subsubsection*{BigBelly Urban Bins} In these experiments, GPT-4o demonstrated the highest performance among the models with an average accuracy of 0.69 across all types. It excelled in classifying electronics (0.9) and glass (0.9), reflecting the model’s capability to manage the more distinct visual features of these materials. However, it encountered significant challenges with cardboard (0.2), largely due to the inability to correctly assess the size of the BigBelly bin openings relative to the size of the items. This limitation led to misclassifications where larger items like cardboard were incorrectly identified as fitting into the bin.

GPT-4o-mini, with an accuracy of 0.51, faced even greater difficulties, particularly with cardboard and wood, where it failed to classify any images correctly (0.0 accuracy for cardboard and 0.3 for wood). This model struggled because it hallucinated the existence of a third bin, often outputting “middle” when only left and right options were available. 

Claude was the weakest performer, achieving an accuracy of 0.47. Similar to GPT-4o-mini, Claude also struggled with larger items like cardboard and wood, failing to recognize that these items were too large for the bin openings. This misjudgment led to frequent misclassifications, further highlighting the model’s limitations in scenarios where precise spatial reasoning and contextual understanding of bin dimensions are critical. 

\subsubsection*{Residential Bins} The performance dynamics shifted slightly. GPT-4o-mini achieved the highest accuracy at 0.73, particularly excelling in the classification of wood (1.0) and maintaining solid performance across other categories like cardboard (0.8) and paper (0.8). However, despite its overall strong performance, GPT-4o-mini also exhibited a surprising issue: it again hallucinated the existence of a third bin on four occasions. This is notable given that, unlike the BigBelly bins where the two bins had three openings that might have caused confusion, the residential bins are clearly separated into only two large and distinct containers. This unexpected behavior suggests that the model may have underlying difficulties in accurately interpreting the physical layout of the bins, even in seemingly straightforward scenarios.

GPT-4o followed closely with an accuracy of 0.63, showing consistent strength in electronics (0.8) and paper (0.8), though its performance in wood (0.7) and cardboard (0.8) was slightly lower than GPT-4o-mini. Claude, on the other hand, showed improvement compared to its performance in the BigBelly bins, achieving an accuracy of 0.64. While GPT-4o-mini’s weakest material in this experiment was electronics (0.2), electronics was one of GPT-4o’s strongest categories.

The higher accuracies across all models in this experiment may not fully reflect the models’ strengths but rather the increased capacity of the 96-gallon residential bins, which can accommodate a broader range of item sizes. This larger bin size likely simplified the classification task, as more items, including larger ones like cardboard and wood, could easily fit into the bins. This suggests that the observed improvement in performance might be partially due to the bin’s capacity rather than solely the models’ predictive capabilities.

\subsubsection*{Three-Category Bins} GPT-4o led with an accuracy of 0.73, demonstrating its ability to effectively manage the complexity of multiple disposal categories. The model excelled in classifying electronics (0.9) and organics (0.9), maintaining high accuracy in textiles (0.9) and glass (0.6).

GPT-4o-mini, with an accuracy of 0.58, showed similar strengths as GPT-4o but struggled significantly more with wood (0.4) and organic matter (0.1). The low accuracy in organic matter classification was particularly concerning, as it stemmed from the model repeatedly outputting “compost” seven times, despite explicit instructions in the prompt to use only “left”, “right”, “middle”, or “none”. This issue highlights a critical weakness in GPT-4o-mini’s ability to follow prompts accurately, especially in complex scenarios involving multiple disposal categories.

Claude, achieving an accuracy of 0.59, showed improvement in this setup compared to simpler bin configurations. However, it continued to face challenges, particularly with wood (0.7) and cardboard (0.3). While Claude’s performance was slightly better here, its ongoing struggles with these materials indicate persistent issues with accurately classifying items, particularly in scenarios where size and material characteristics are critical.
Across the different models, GPT-4o consistently delivered the best results. However, all models faced challenges, particularly with accurately classifying items based on their size, which was a recurring issue across different materials like cardboard and wood. This underscores the need for further refinement, especially in handling larger or less distinct items and ensuring that models adhere strictly to the provided prompts.

\begin{figure}
  \centering
  \includegraphics[width=5.5in]{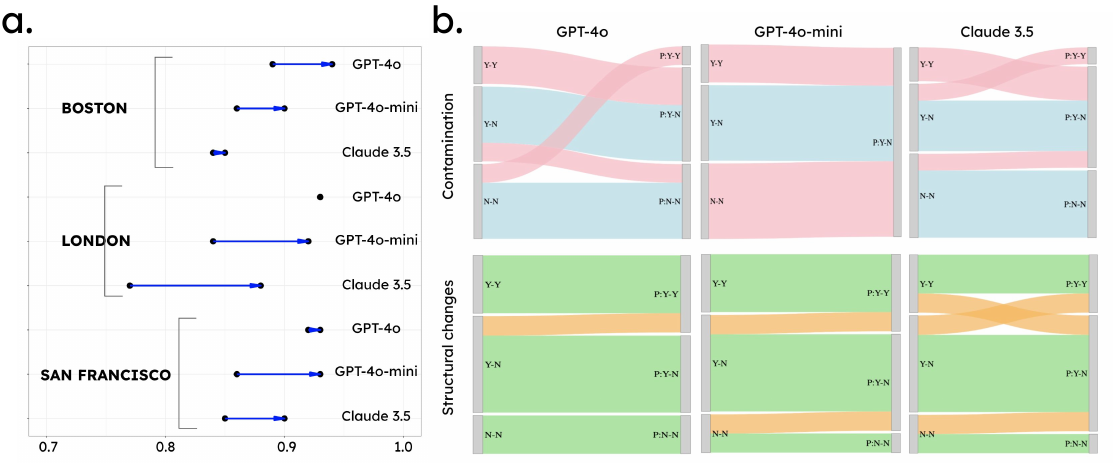}
  \caption{Three additional tests for contextual predictions. (a) City-specific guidelines were added to the prompt. The direction of the arrows indicates that the classification improved (or remained the same) as a result. (b) Pairs of images were tested for contamination (e.g., soiled item) and structural changes (e.g., broken glass). The changes in the pair predictions are illustrated with the blue and pink bands corresponding to correct and incorrect items, respectively. For instance, for GPT-4o, four pairs were Y-N pairs (recyclable prior to contamination / not recyclable after contamination); three items were predicted correctly (blue) but one item was predicted to be N-N. In general, contaminated items that are still recyclable were incorrectly predicted as no longer being recyclable.}
  \label{exp2-4}
\end{figure}

\subsection{Experiment 2: Location-Specific Guidelines (Figure 3)}
Here, we tested the models’ performance under different location-specific guidelines: Boston, London, and San Francisco. Each model was evaluated using the same set of 100 images across all locations to assess how well they could adapt to varying recycling guidelines.

When comparing the “No Location” scenario (which used Boston labels) to the Boston-specific prompt, there was a noticeable improvement in accuracy for all models. GPT-4o’s accuracy increased from 0.89 in the “No Location” scenario to 0.94 with the Boston guidelines. GPT-4o-mini also saw a jump from 0.86 to 0.90, while Claude’s accuracy slightly increased from 0.84 to 0.85. These improvements suggest that the models benefitted from the more explicit instructions provided by the Boston-specific guidelines, even though the prompt specified Boston as the city of the “No Location” scenario.
Across all locations, only five items changed their recyclability status when switching from Boston Regulations to London or San Francisco’s Regulations. This limited variation implies that the increase in accuracy seen in location-specific scenarios may not be entirely due to the models adapting to new guidelines, but rather to the added support of a detailed prompt. The consistency in the incorrect images also supports this, as many images were incorrect regardless of the location-specific prompt.

Overall, the results indicate that while the models are capable of adjusting to different location-specific guidelines, the relatively minor changes in recyclability in this dataset across locations suggest that the observed improvements in accuracy may be attributable to the clarity and specificity of the guidelines rather than the models’ ability to handle different regional recycling practices.

\subsection{Experiment 3: Structural Changes}
The models were evaluated for their ability to classify items before and after undergoing physical transformations due to contamination or structural changes. GPT-4o achieved an overall combined image-pair accuracy of 0.75, and a slightly higher individual image accuracy of 0.875. 

For contamination, GPT-4o led with a combined accuracy of 0.6 and an individual accuracy of 0.8, demonstrating a solid performance in recognizing and adapting to soiled items. GPT-4o-mini followed with a combined accuracy of 0.4, showing some limitations in handling subtle contamination, while Claude achieved a combined accuracy of 0.7.

In the structural changes scenario, GPT-4o again excelled with a combined accuracy of 0.9 and an individual accuracy of 0.95, showing strong adaptability to physical alterations. GPT-4o-mini achieved a combined accuracy of 0.8, and Claude lagged behind with 0.5, indicating difficulties in accurately reclassifying structurally altered items. These results highlight GPT-4o’s robustness in managing complex real-world scenarios, though further refinement is needed across all models to enhance their practical applicability in diverse recycling contexts.

\subsection{Experiment 4: Multi-Material Objects}
This experiment tested the models’ ability to classify items composed of multiple materials, such as plastic combined with metal or glass. GPT-4o led with an impressive overall accuracy of 0.98, misclassifying only 1 out of 50 images. This high accuracy indicates GPT-4o’s strong capability in managing the complexity of multi-material objects, effectively identifying and categorizing them with minimal errors.

GPT-4o-mini also performed well, achieving an overall accuracy of 0.92, although it misclassified 4 out of 50 images. This slightly lower accuracy compared to GPT-4o suggests that while GPT-4o-mini is competent, it still has room for improvement in handling the intricacies of multi-material items.
Claude, however, demonstrated significant challenges in this experiment, with an overall accuracy of 0.78 and 11 misclassified images out of 50. The lower accuracy indicates that Claude struggled with the complexities of multi-material objects, particularly when compared to the more robust performance of GPT-4o.

In summary, GPT-4o was the most reliable model for classifying multi-material objects, showing a clear advantage over GPT-4o-mini and especially Claude. The consistent performance of GPT-4o across these complex scenarios highlights its effectiveness, while the noticeable gap in accuracy for Claude underscores the need for further refinement in handling multi-material classifications.

\section{Discussion}

Standard guidelines for recycling are often inadequate for the public in determining the recyclability of an item due the myriad of its possible conditions. As we have shown here, the rapid advances in vision-language models now offer contextual intelligence that leads to exceptional performance in many complex situations. GPT-4o, for instance, excels in many of the tests carried out in this work.

A continuing challenging in evaluating the prediction models is the difficulty of ascertaining the true label, especially given the more than dozen disposal methods specified in most city guidelines and their variations. We have called, emailed, and visited city recycling managers to obtain the true labels for many contexts. In some cases, the answers were clear (e.g., a greasy pizza box is recyclable in Boston but not in some other cities); but in other cases, there were no definitive answers, in part because the different companies who buy the recycled items at a given time have different requirements. Nevertheless, the models we tested here are ideal for their easy-to-modify prompts in addition to their contextual intelligence.

In our evaluation, no distinction was between a false positive and a false negative prediction. In reality, a false positive (an item predicted to be recyclable but is not) is more costly, as it contaminates the batch to which it was assigned. A weight function could be incorporated into the model to improve its practicality.

\bibliography{reference}

\end{document}